# COMPLEX SCENE CLASSIFICATION OF POLSAR IMAGERY BASED ON A SELF-PACED LEARNING APPROACH

Wenshuai Chen[1], Shuiping Gou[1], Member, IEEE, Xinlin Wang[1], Licheng Jiao[1], Fellow, IEEE, Changzhe Jiao[1], Member, IEEE, Alina Zare[2], Senior Member, IEEE

[1] Key Laboratory of Intelligent Perception and Image Understanding, Ministry of Education of China, School of Artificial Intelligence, Xidian University, Xi'an, China, 710071
[2] Department of Electrical and Computer Engineering, University of Florida, Gainesville, FL 32611, USA

*Abstract* – Existing polarimetric synthetic aperture radar (PolSAR) image classification methods cannot achieve satisfactory performance on complex scenes characterized by several types of land cover with significant levels of noise or similar scattering properties across land cover types. Hence, we propose a supervised classification method aimed at constructing a classifier based on self-paced learning (SPL). SPL has been demonstrated to be effective at dealing with complex data while providing classifier.

In this paper, a novel Support Vector Machine (SVM) algorithm based on SPL with neighborhood constraints (SVM_SPLNC) is proposed. The proposed method leverages the easiest samples first to obtain an initial parameter vector. Then, more complex samples are gradually incorporated to update the parameter vector iteratively. Moreover, neighborhood constraints are introduced during the training process to further improve performance. Experimental results on three real PolSAR images show that the proposed method performs well on complex scenes.

## I. INTRODUCTION

Polarimetric synthetic aperture radar (PolSAR) technology plays an important role in military, agriculture, geology and other application areas [1], [2]. Due to this, PolSAR image classification is a very active area of research where many effective methods have been proposed. Extracting polarization features and designing an appropriate classifier are common steps in these methods. Polarization features have been extracted using several decomposition techniques [3] including Pauli decomposition [4], Cloude-Pottier decomposition [5], and Freeman decomposition [6].

After polarization feature extraction, it is important to design an appropriate classifier. Studies have reported results using the support vector machine (SVM) [7], random forest [8], artificial neural networks [9], and other machine learning methods [10]. Another common classification method in the PolSAR literature is the Wishart classifier (WC) [11-13]. The above methods can classify terrain given PolSAR imagery. However, for more complex scenes characterized by several types of land cover with similar scattering properties, scenes with mixed scattering mechanisms, and scenes with significant levels of noise, these methods do not provide reliable performance.

Recently, due to the significant success of deep learning across a range of applications and data, classification of PolSAR images using deep learning architectures have been investigated [14-16]. Deep neural networks (DNNs) architectures perform well because of their strong non-linear fitting ability. Nonetheless, DNNs have a large number of parameters and, thus, require large training data sets to optimize these parameters. Training DNNs is also a time- and resource-consuming process. In addition, obtaining the labels of the PolSAR data for network training is time-consuming.

Self-paced learning (SPL) has attracted increased attention in recent years [17, 21]. SPL has been widely used in many problems including specific-class segmentation [18], long-term tracking [19] and visual category discovery [20]. In SAR image processing, Shang et al. [22] proposed an algorithm based on SPL for change detection in SAR imagery which outperforms state-of-the-art algorithms in terms of accuracy.

SPL has shown excellent performance in a wide range of classification problems. SPL's learning mechanism is inspired by the human learning process in which the easiest aspects of a task learned first and, then, more difficult aspects are incorporated and learned. This learning mechanism has been empirically demonstrated to be robust to noisy data and be instrumental in avoiding local minima to achieve better generalization results [23-25].

In this paper, we propose the use of SPL within a novel SVM algorithm for classification of complex PolSAR scenes. Furthermore, we introduce a new self-paced regularization term that incorporates spatial information. Considering the robustness and non-linear fitting ability of SVM algorithm, a novel SVM algorithm based on SPL with Neighborhood Constraints (SVM_SPLNC) for PolSAR image classification is proposed. Under this learning mechanism, the SVM algorithm learns the easier samples first and then gradually involves more difficult samples in the training process.

This paper is organized as follows. Section II introduces the concept of SPL. Section III describes the proposed novel svm with self-paced learning and neighborhood constraints (SVM_SPLNC) method. Experiment results on three measured PolSAR images are reported in Section IV. Finally, conclusions and potential future work are discussed.

## II. Self-Paced Learning Background

As opposed to traditional machine learning methods which consider all samples simultaneously, SPL presents training data in a meaningful order to facilitate learning. The order of the samples is determined by their learning difficulty. However, a key issue with this approach is that, typically, we are not provided with a readily computable measure of the learning difficulty associated with each sample. We address this issue using the concept of a loss function. A well-accepted assumption is that the smaller the training loss of a sample, the more likely it is to be an "easy" sample. Therefore, the relationship between the learning difficulties of sample and the training loss can be established based on this assumption. In SPL, a weight $v$ between 0 and 1 is used to denote the learning difficulties of samples, and a gradually increasing pace parameter $\lambda$ is introduced to control the pace for learning new samples. The value of $v$ is determined by a regularization term $f(v,\lambda)$ called "self-pace regularization term". The model of SPL is formally elaborated below.

Given a training dataset $D = \{(x_i, y_i), i=1...n\}$, in which $x_i$ denotes the $i^{th}$ observed sample, and $y_i$ represents its label, let $L_i = L(y_i, g(x_i, w))$ denote the loss of sample $x_i$, which is the cost between the label $y_i$ and the estimated label from the classifier, $g(x_i, w)$. Here $w$ represents the model parameters of the classifier. Furthermore, let $p(w)$ be a regularization term imposed on classifier parameters. A general regularized machine learning objective function can then be expressed as shown in (1),

$$w = \arg\min_{w} \sum_{i=1}^{n} L(y_i, g(x_i, w)) + p(w) \qquad (1)$$

In SPL, a weight $v$ between 0 and 1 is used to denote the learning difficulty of a sample where a small weight corresponds to more difficult samples. Also, a pace parameter $\lambda$ is used to control the pace for learning new samples. The value of $v$ is determined by a regularization term $f(v,\lambda)$ called the *self-paced regularization term*. Thus, the SPL model differs from (1) by using a weighted loss term $v_i L(y_i, g(x_i, w))$ for each sample where $v_i \in [0,1]$, $i=1...n$, and incorporating the self-paced regularization term $f(v_i, \lambda)$ imposed on sample weights,

$$(w, v) = \arg\min_{w,v} \sum_{i=1}^{n} v_i L(y_i, g(x_i, w)) + p(w) + f(v_i, \lambda), \qquad (2)$$

The weight vector $v = [v_1...v_i...v_n]$ is defined based on the following two rules [26]:

a. $v_i$ is monotonically decreasing with respect to increasing training loss such that $\lim_{L_i \to 0} v_i = 1$ and $\lim_{L_i \to \infty} v_i = 0$

b. $v_i$ is monotonically increasing with respect to the pace parameter $\lambda$, such that $\lim_{\lambda \to 0} v_i = 0$ and $\lim_{\lambda \to \infty} v_i = 1$.

Rule (a) indicates that the model more heavily weights easy samples with smaller training loss. Rule (b) indicates that as the pace parameter $\lambda$ gets larger, the model increases the weight of all samples, thus incorporating more complex samples into the training procedure. Given these axiomatic rules, Meng et al. [26] proposed two self-paced regularization terms: the binary and the linear regularization terms.

1. Binary regularization term

The binary self-paced regularization term can be expressed as:

$$f(v_i, \lambda) = -\lambda v_i \qquad (3)$$

subject to the constraint that the weight of each sample is binary, i.e., $v_i \in \{0,1\}$.

When plugging (3) into the SPL regularization term in (2) and simplifying the equation, it can be seen that $v_i$ can be

obtained by:

$$v_i = \arg\min_v \sum_{i=1}^n v_i L(y_i, g(x_i, w)) + p(w) - \lambda v_i$$
$$= \arg\min_v \sum_{i=1}^n v_i (L_i - \lambda) + p(w) \quad (4)$$
$$\Rightarrow \begin{cases} v_i = 1 & if\ L_i < \lambda \\ v_i = 0 & if\ L_i \geq \lambda \end{cases}$$

When the weight of the $i^{th}$ sample $v_i$ is 1 in an iteration, the sample is considered to an easy sample and it will be used during learning.

---

**Algorithm 1.** SPL Training Procedure

**Input:** Training set $D = \{(x_i, y_i),\ i=1...n\}$

**Output:** Model parameters $w$

**SPL Training Procedure:**

Step1: Initialize weights of all samples $v = [v_1...v_n]$ and parameter $\lambda$.

Step2: Fix $v$, update $w$ using (2).

Step3: Fix $w$, calculate the training loss $L(y_i, g(x_i, w))$ for each sample, then update $v$ using (4) or (6).

Step4: Update $\lambda$ using $\lambda = \kappa\lambda\ \kappa > 1$.

Step5: Repeat steps 2-4 until the mean of all $v_i$ values is equal to or approximately equal to 1. Return the estimated $w$ parameters.

---

2. Linear regularization term

The linear self-paced regularization term is shown in (5):

$$f(v_i, \lambda) = \lambda\left(\frac{1}{2}v_i^2 - v_i\right) \quad (5)$$

With (5), (2) and simplifying the equation, $v_i$ can be obtained by:

$$v_i = \arg\min_v \sum_{i=1}^n v_i L(y_i, g(x_i, w)) + p(w) - \lambda\left(\frac{1}{2}v_i^2 - v_i\right)$$
$$= \arg\min_v \sum_{i=1}^n v_i \left(\frac{1}{2}\lambda v_i + L_i - \lambda\right) + p(w) \quad (6)$$

$$\begin{cases} v_i = 1 - \dfrac{L_i}{\lambda} & if\ L_i < \lambda \\ v_i = 0 & if\ L_i \geq \lambda \end{cases} \quad (7)$$

Under the linear regularization term, when the training loss $L_i$ of the $i^{th}$ sample is less than the pace parameter $\lambda$, the weight of this sample is a value between 0 and 1.

In SPL, the parameter vector, **w**, and the sample weights, **v**, are updated iteratively with the procedure outlined in Alg. 1.

### III. PROPOSED METHOD

Support Vector Machines, given their ability address non-linearity and high generalization capabilities, have potential for the classification of complex PolSAR imagery [27]. Thus, we extend the SVM with SPL using a neighborhood constrained self-paced regularization function.

In our approach features are extracted using matrix and Cloude-Pottier decomposition from PolSAR image[4]. Specifically, the first three eigenvalues of the coherency matrix and their sum are used as feaures. Furthermore, the entropy, alpha and anisotropy parameters of coherency matrix T are also used as features.

In our approach, each pixel is represented by a row feature vector, and the feature vectors are obtained by matrix decomposition and Cloude-Pottier decomposition [4]. The eigenvalues $\lambda_1, \lambda_2, \lambda_3$ of the coherency matrix T and their sum are taken as features. In addition, the Entropy, Alpha and Anisotropy parameters are also taken as features. The SVM presents high generalization capability even in linearly non-separable circumstances, they have a great potential for the classification of PolSAR images [27]. The SPL is used to improve the performances of SVM classifier on the complex scenes. In addition, the neighborhood information of each pixel in the image is beneficial to improve the classification accuracy. Therefore, a new SPL regularization term with neighborhood constraints is designed.

#### A. Proposed Model

When training an SVM, the goal is to maximize the margin between two classes while maintaining high classification accuracy. This goal is defined mathematically in the SVM objective function:

$$(w, b) = \arg\min_{w,b} \sum_{i=1}^n L(y_i, (wx_i + b)) + c\|w\|^2. \quad (8)$$

where $\{w, b\}$ denotes the classifier parameters, which includes a coefficient vector $w$ and a bias term $b$, and $L(y_i, (wx_i + b))$ is the hinge loss function calculated as:

$$L(y_i, (wx_i + b)) = \max(0, 1 - y_i(w^T x_i + b)) \quad (9).$$

The parameter $c$ is the standard regularization parameter trading off between the hinge loss and the margin size. A large $c$ will lead to the larger margin between the classes (and, potentially, increased error rates). Hence, the model will have lower error rates (and the potential to overfit) with a small $c$ value.

To incorporate self-paced learning into the SVM training procedure, weighted samples are used to learn the parameter vector at each iteration, and the self-paced regularization term is incorporated into the objective function. Therefore, the proposed model SVM_SPLNC can be formulated as:

$$(\mathbf{w},b,\mathbf{v}) = \arg\min_{\mathbf{w},b,\mathbf{v}} \sum_{i=1}^{n} v_i L(y_i,(\mathbf{w}\mathbf{x}_i+b)) + c\|\mathbf{w}\|^2 + f'(v_i,\lambda). \quad (10)$$

When $v_i = 0$, the loss incurred by the $i^{th}$ sample is always zero and when the $v_i$ values for all samples are equal to 1, (10) collapses to the conventional SVM objective shown in (8). The pace parameter $\lambda$ controls the learning process, with parameter $c$ trading off the margin and other items. Both $c$ and $\lambda$ are initialized before training.

$f'(v_i,\lambda)$ is the new self-paced regularization term with neighborhood constraints imposed on the weight $v_i$. In PolSAR imagery, each pixel generally has physical properties in common with pixels in its spatial neighborhood. Therefore, using neighborhood information can help to improve classification accuracy [26,33]. In order to incorporate this information, the weight $v_i$ for a data point is determined using a linear combination between the training loss for the data point under consideration as well as the average loss for the sample's spatial neighborhood. Thus, (7) can be rewritten as follows:

$$\begin{cases} v_i = 1 - \dfrac{L_i + \gamma_i L_i'}{\lambda} & \text{if } L_i + \gamma_i L_i' < \lambda \\ v_i = \quad 0 & \text{if } L_i + \gamma_i L_i' \geq \lambda \end{cases} \quad (11)$$

where $L_i' = \dfrac{1}{8}\sum_{j=1}^{8} L_{ij}$, $p_{ij} = \dfrac{L_{ij}}{L_i'}$, $\gamma_i = -\sum_{j} p_{ij}\log(p_{ij})$

$L_i$ denotes the training loss of the $i^{th}$ sample and $L_i'$ represents the average training loss of the $i^{th}$ sample's eight neighboring pixels $L_{ij}$. The parameter $\gamma_i$ is used to tradeoff between $L_i$ and $L_i'$. The value for the $\gamma_i$ parameter can be determined by the Shannon entropy of neighborhood pixels' training loss. Namely, the larger the Shannon entropy, the more likely the $i^{th}$ sample belongs to a homogeneous region and $L_i$ is subject to a stronger neighborhood constraints. The value $\dfrac{L_i + \gamma_i L_i'}{1+\gamma_i}$ represents the training loss of the central pixel. Fig.1 illustrates the $i^{th}$ sample's training loss with eight neighborhood constraints.

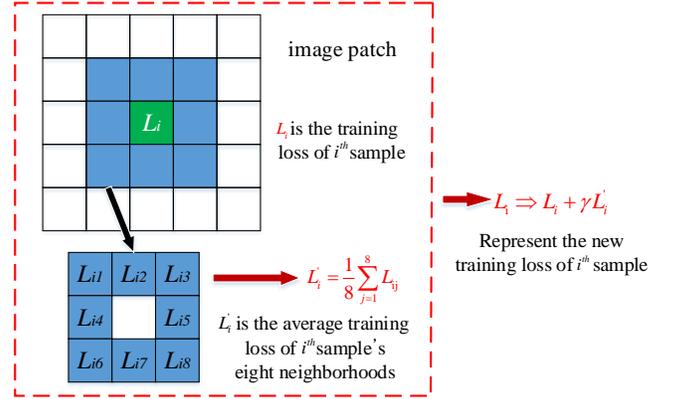

Fig.1. $i^{th}$ sample's training loss with eight neighborhood constraints

B. *Solving the Proposed Model*

Eqn. (10) can be rewritten as:

$$\begin{cases} (\mathbf{w},b) = \arg\min_{\mathbf{w},b,\mathbf{v}} \dfrac{1}{2}\|\mathbf{w}\|^2 + c\sum_{i=1}^{n} v_i l_i + f'(v_i,\lambda) \\ s.t. \ y_i(\mathbf{w}^T\mathbf{x}_i + b) \geq 1 - l_i, \ l_i \geq 0, \ v_i \in [0,1] \end{cases} \quad (12)$$

Eq. (12) is optimized iteratively as outlined below.

*Step 1. Initialization*

Initialize the weight $\mathbf{v} = [v_1...v_n]$, $v_i$ is randomly given to positive numbers, approach to 0. $\lambda$ is given a number, $c$, is kernel function.

*Step 2. Fix $\mathbf{v}$ and optimize $\mathbf{w}$ and $b$.*

Given fixed $\mathbf{v}$ values, the objective function can be expressed as:

$$\begin{cases} (\mathbf{w},b) = \arg\min_{\mathbf{w},b} \dfrac{1}{2}\|\mathbf{w}\|^2 + c\sum_{i=1}^{n} v_i l_i \\ s.t. \ y_i(\mathbf{w}^T\mathbf{x}_i + b) \geq 1 - l_i \ \ l_i \geq 0 \end{cases} \quad (13)$$

With this form, the optimization process of $\mathbf{w}$ and $b$ is similar to the process used for a conventional SVM. Introducing Lagrange multipliers $\beta$ and $\delta$, the Lagrangian of the problem can be defined as:

$$\mathrm{T} = \min_{\delta>0,\beta>0} \dfrac{1}{2}\|\mathbf{w}\|^2 + c\sum_{i=1}^{n} v_i l_i + \\ \sum_{i=1}^{n}\delta_i(1 - l_i - y_i(\mathbf{w}^T x_i + b)) - \sum_{i=1}^{n}\beta_i l_i \quad (14)$$

According to the Karush-Kuhn-Tucker (KKT) conditions, the optimal solution must satisfy the conditions listed in (14):

$$\begin{cases} \dfrac{\partial T}{\partial \mathbf{w}} = \mathbf{w} - \sum_{i=1}^{n} \delta_i y_i \mathbf{x}_i = 0, \\ \dfrac{\partial T}{\partial b} = \sum_{i=1}^{n} \delta_i y_i = 0 \\ \dfrac{\partial T}{\partial l_i} = cv_i - \delta_i - \beta_i = 0 \end{cases} \quad (15)$$

Substituting (15) into (14), the objective function (13) is modified to the following dual form:

$$\begin{cases} \max_{\boldsymbol{\delta}} \sum_{i=1}^{n} \delta_i - \dfrac{1}{2} \sum_{i=1}^{n} \sum_{j=1}^{n} \delta_i \delta_j y_i y_j K(\mathbf{x}_i, \mathbf{x}_j) \\ s.t. \sum_{i=1}^{n} y_i \delta_i = 0, \ 0 \leq \delta_i \leq cv_i \end{cases} \quad (16)$$

where $\delta_i$ is [0, $cv_i$]. The $K(\mathbf{x}_i, \mathbf{x}_j)$ represent the kernel function. Gaussian kernel is selected as kernel function [29]. Since Eq. (16) is a quadratic programming in its dual form, we used CVX toolkits [30] to solve it.

*Step 3. Fix **w** and b, then optimize **v***

After obtaining $\boldsymbol{\delta}$, the value of **w**, b can be calculated using (14), and the value of $l_i$ by (9). Then, given fixed w and b values the objective function in (11) becomes:

$$\mathbf{v} = \min_{\mathbf{v}} \ c \sum_{i=1}^{n} v_i l_i + f'(v_i, \lambda) \quad (17)$$

where $v_i$ is calculated using (10).

*Step 4. Update $\lambda$, then repeat steps 2-3*

After updating the SVM parameters and the SPL weight for each data point, the $\lambda$ value is updated according to (18),

$$\lambda = \kappa \lambda, \ \kappa > 1 \quad (18)$$

where k is constant of the step update.

Then, Step 2 and Step 3 are repeated until the mean of **v** is equal to 1. All training samples have been included in the training set used to update the SVM.

## IV. EXPERIMENTS

In this section, three measured PolSAR data sets are used to validate the performances of the proposed method. The proposed method is compared with three typical PolSAR classification methods, including the SVM [31], the Wishart classifier (WC) [32] and Sparse Representation-based classification (SRC) [33]. For the SVM, a radial basis function (RBF) kernel is used, the parameter gamma for RBF is 1, and the tolerance of termination criterion and the cost factor are 0.00001 and 50. For WC, the training samples are used to calculate the Wishart centers of each class, and the Wishart distance is used to classify each pixel. For SRC, an over-complete dictionary is first generated from the training samples. Then, the test samples are classified by obtaining the sparse representation of the test samples and calculating the residuals of each class. Finally, in order to validate the newly constructed regularization term, the SPL based SVM with original linear regularization term (SVMSPL) is also used as a comparison algorithm.

For the proposed SVM_SPLNC algorithm, the parameters are set as the RBF kernel, $c$=100, $\gamma=1$, $\lambda=0.1$, $k$=1.05 which are based on our experimental experience. For these comparison algorithms, we use the same numbers of training samples per class and extract the same. All the experiments are performed on an Intel i5-6500 CPU 3.2GHz, and the code is written with MATLAB R2015b development environment.

### A. Flevoland Data Set from AIRSAR

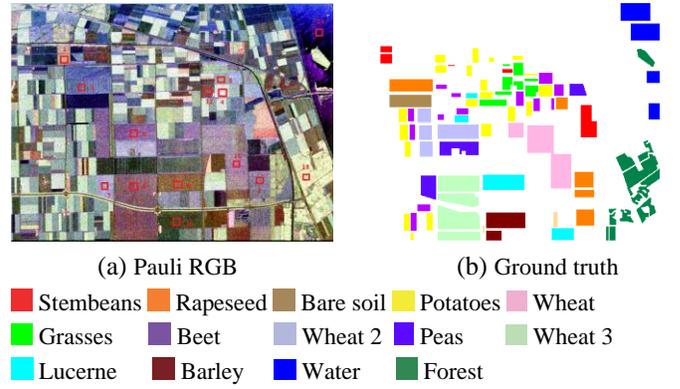

(a) Pauli RGB          (b) Ground truth

■ Stembeans ■ Rapeseed ■ Bare soil ■ Potatoes ■ Wheat
■ Grasses ■ Beet ■ Wheat 2 ■ Peas ■ Wheat 3
■ Lucerne ■ Barley ■ Water ■ Forest

Fig.2. (a) Pauli RGB of Flevoland. (b) Ground truth

The first data set investigated was the NASA/JPL AIRSAR L-Band four-look PolSAR data of Flevoland, Netherlands, which has the size of $1024 \times 750$ pixels. The spatial resolution is 6.6 m in the slant range direction and 12.10 m in the azimuth direction. Fig. 2 illustrates the corresponding Pauli RGB image and ground truth, respectively. The ground-truth map was obtained from reference [35]. There are 14 classes in this data with each class indicating a type of land covering. A total of 167712 pixels are labeled as ground truth and only 2% of them were used to train the classifiers with the remainder used for testing. The training areas are marked with red blocks in Fig. 2(a), which are randomly selected from each terrains. The reported testing accuracies are obtained by testing on the 98% residual pixels. For the SVM_SPLNC, the parameters are follows $c$=100, $\lambda=0.1$, $k$=1.05 and kernel function is RBF.

**a.** Convergence analysis

Fig. 3 shows that the average training loss over all of the training points each iteration for multiple runs is decreases during the training process, indicating the proposed algorithm is likely to converge. In the experiments, the weights of all samples were randomly initialized positive numbers approaching 0, then updated at each iteration. We can see the weight increases with the number of iterations in Fig. 3. At iteration 35, the average weight is only 0.5, indicating that all training samples have not been included or learned. Given that only a portion of the samples have been learned, the generalization and classification ability of the model at iteration 35 is poor. For example, the overall accuracy is 0.66 at this iteration as shown in Fig. 4(b) where several easy samples (e.g., Forest, Lucerne, and Bare soil) are classified correctly but other complex samples (such as Wheat, Wheat 3 and Rapeseed, as shown with the red circles) are misclassified. Fig. 4(c) shows the overall accuracy is 0.81 at 50 iterations of the model, and the average weight for training samples is 0.7. The average weight is close to 1 at iteration 90. In this iteration, it can be considered that the entire training data has been included in the training process. The final result is relatively very good with an overall accuracy of 0.91 as shown in Fig. 4(d) with Wheat, Wheat 3 and Rapeseed being correctly classified (as shown in the red circles).

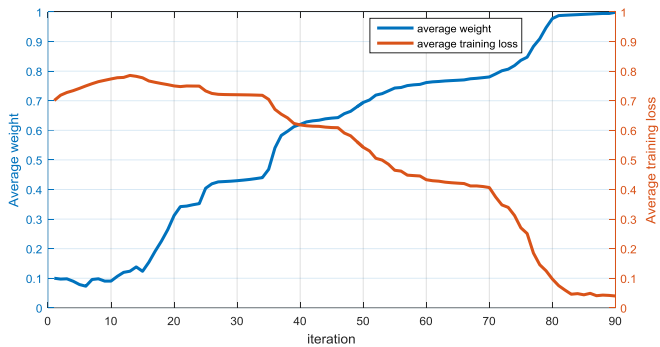

Fig. 3 average weight and average training loss of all training samples vary with the number of iterations in the process of training respectively.

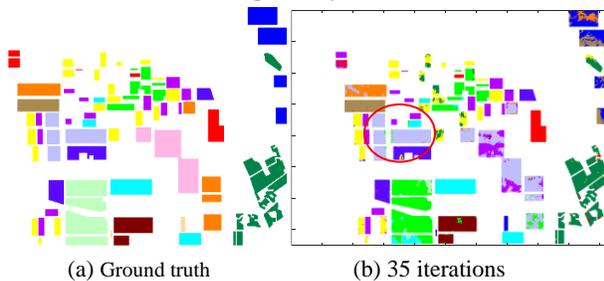

(a) Ground truth        (b) 35 iterations

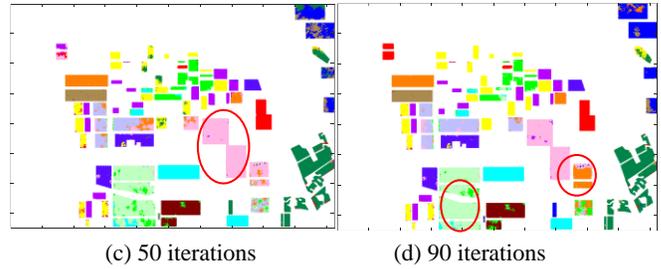

(c) 50 iterations        (d) 90 iterations

Fig.4 (a) Ground truth of Flevoland. (b)~(d) classification results of models obtained at different number of iterations

**b.** Classification results

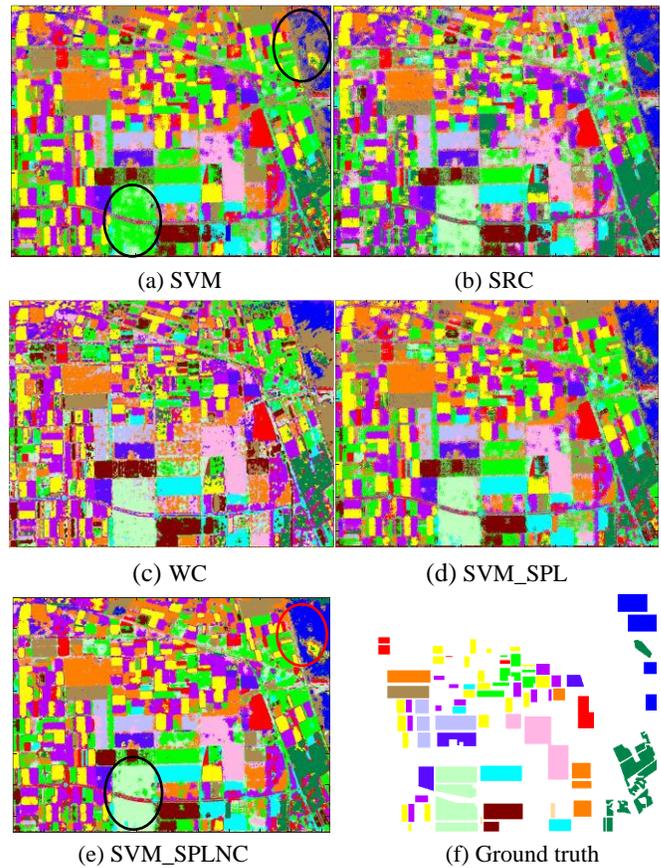

(a) SVM        (b) SRC

(c) WC        (d) SVM_SPL

(e) SVM_SPLNC        (f) Ground truth

Fig.5. classification results of five methods

Fig. 5 shows visual classification results with accuracies for each class are listed in Table I. To compare the SVM and the proposed SVM_SPLNC method, the two methods are implemented using CVX toolkit [30]. As shown with black circles in Fig. 5(a), the SVM cannot distinguish Rapeseed, Potatoes, Wheat 3 and Water from each other very well. For the SRC method, results indicated that most of the Rapeseed, Potatoes and Grasses were misclassified to other categories. WC from [35] (see Fig. 5(c)) had satisfactory accuracies on most categories except for Water and Rapeseed. The proposed SVM_SPLNC method had the highest overall accuracy of 0.91. Furthermore, from Fig. 5 we can see that the classification result by SPLNC is smoother and outperforms comparison methods on most of the classes including Rapeseed, Potatoes, Wheat 2, Wheat 3 and Water. Compared to SVM, our method trained with the learning mechanism of SPL produces better results, e.g., Wheat 3 and Water are recognized better (see black circles in Fig 5.(a) and (e)), and the overall accuracy (OA) increases by 10%. Moreover, compared to SVM_SPL, SVM_SPLNC has higher accuracy rate for most land cover types indicating that the proposed regularization term with neighborhood constraints is effective.

TABLE I
ACCURACIES OF FLEVOLAND DATA SET FROM AIRSAR
AA: AVERAGE ACCURACY; OA: OVERALL ACCURACY

|  | *SVM* | *SRC* | *WC* | *SVM_SPL* | *SVM_SPLNC* |
|---|---|---|---|---|---|
| *Stembeans* | 0.9721 | 0.9642 | 0.9508 | 0.9615 | **0.9874** |
| *Rapeseed* | 0.7175 | 0.6049 | 0.7484 | 0.7617 | **0.7902** |
| *Bare soil* | 0.9933 | 0.9211 | 0.9920 | 0.9702 | **0.9975** |
| *Potatoes* | 0.9800 | 0.6631 | 0.8775 | 0.9638 | **0.9865** |
| *Beet* | 0.9540 | 0.9561 | 0.9513 | 0.9549 | **0.9788** |
| *Wheat 2* | 0.7323 | 0.7797 | 0.8272 | 0.7945 | **0.8393** |
| *Peas* | 0.9259 | 0.9396 | **0.9628** | 0.8968 | 0.9428 |
| *Wheat 3* | 0.2460 | 0.8226 | 0.8864 | 0.9049 | **0.9277** |
| *Lucerne* | 0.9293 | 0.9513 | 0.9293 | 0.9733 | **0.9760** |
| *Barley* | 0.9329 | 0.9322 | **0.9526** | 0.9476 | 0.8822 |
| *Wheat* | 0.8313 | 0.7610 | 0.8622 | 0.8367 | **0.8605** |
| *Grasses* | **0.9289** | 0.6284 | 0.7246 | 0.7516 | 0.8113 |
| *Forest* | 0.7891 | **0.9797** | 0.8791 | 0.9021 | 0.9239 |
| *Water* | 0.4263 | 0.8002 | 0.5175 | 0.7666 | **0.8435** |
| *AA* | 0.8113 | 0.8360 | 0.8616 | 0.8847 | **0.9105** |
| *OA* | 0.7528 | 0.8231 | 0.8504 | 0.8797 | **0.9067** |

*B. San Francisco Data Set*

The San Francisco data set is a fully polarimetric L-band airborne SAR data set acquired by the AIRSAR sensor of the NASA/JPL, which has the size of 1024×900 pixels. The scene is comprised of Urban, Vegetation, Mountain, Ocean and Bare soil classes. In Fig. 6(a), the data are represented as RGB color composed of the Pauli matrix representation. In this experiment, 0.5% samples (the areas with red blocks in Fig. 6(a)) are used to train the classifier. For the algorithm SVM_SPLNC, we set parameters as RBF kernel, $c$=50, $gamma$=1, $\lambda=0.1$, $k$=1.1 based on this experimental data.

Fig. 6(b)~(f) shows the visual classification results of the respective algorithms. The SVM confuses the ocean class with the orban class in the top right area of the image. SRC has the ability to recognize the ocean class but the results of other categories contain a significant amount of noise. The Wishart classifier misclassifies nearly half of the ocean class in the bottom left area as bare soil. The mountain class in the top left area of the image is more difficult to classify because the shadows in this area are easily misinterpreted as the urban class. Fig. 6(e)-(f) show better results in this area, which correspond to SVM_SPL and SVM_SPLNC.

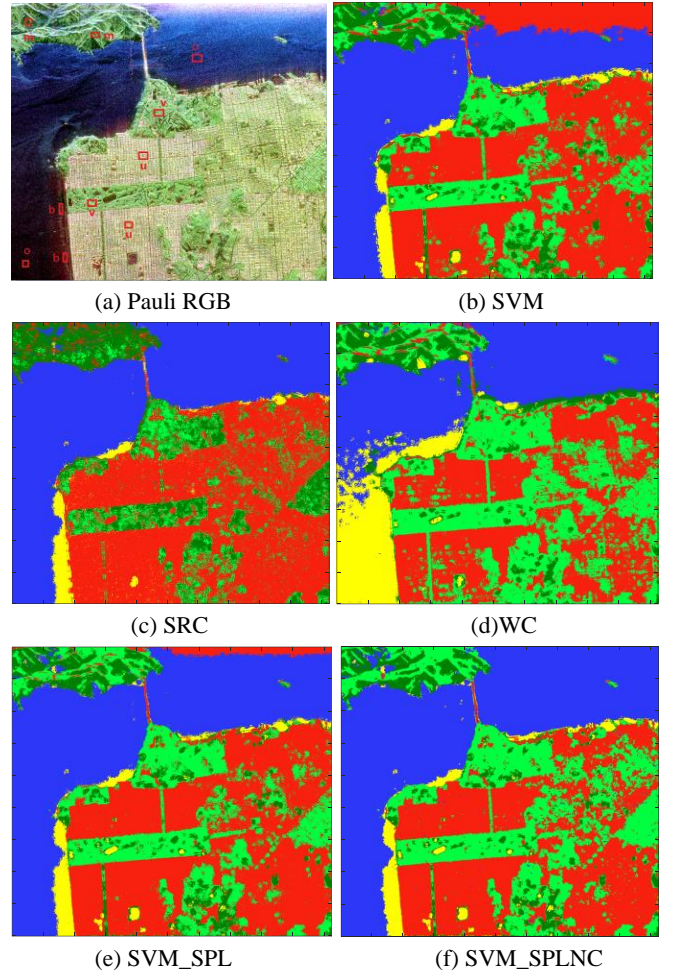

(a) Pauli RGB      (b) SVM
(c) SRC      (d) WC
(e) SVM_SPL      (f) SVM_SPLNC

■ Ocean ■ Vegetation ■ Urban area ■ Mountain ■ Bare soil

Fig.6. (a) Pauli RGB of San Francisco. (b)~(f) classification results of SVM, SRC, WC, SVM_SPL and SVM_SPLNC

*C. Flevoland Data Set from RADARSAT-2*

The Flevoland data set from RADARSAT-2 is a C-band single-look fully PolSAR data with a resolution of 10×5 m and was obtained at fine quad-mode in 2008. A sub-region of 1200 × 1400 pixels was selected, as shown in Fig. 7(a). The areas with red blocks are used to train the classifier and corresponds to 0.4% of the data. The ground-truth reference map is shown in Fig. 7(b) [36]. There are mainly four types of terrain: 1) forest; 2) cropland; 3) water; and 4) urban area. For the algorithm SVM_SPLNC, we set parameters as RBF kernel, $c$=30, $gamma$=5, $\lambda=0.1$, $k$=1.1 in this experiment.

Fig. 8 shows visual classification results for the methods, and Table II lists the classification accuracies obtained for each class. The proposed SVM_SPLNC method provided the best visual results and highest overall accuracy at 0.91. The improvement is obtained primarily from the better performance on the urban areas. Urban areas have a mixed scattering mechanism that results in areas that are more difficult to classify than the other categories. SVM_SPLNC performs better than SVM and SRC in recognizing these complex terrains. Although the proposed SVM_SPLNC performs worse than SRC in recognizing Water and Forest, it still provides a better overall accuracy.

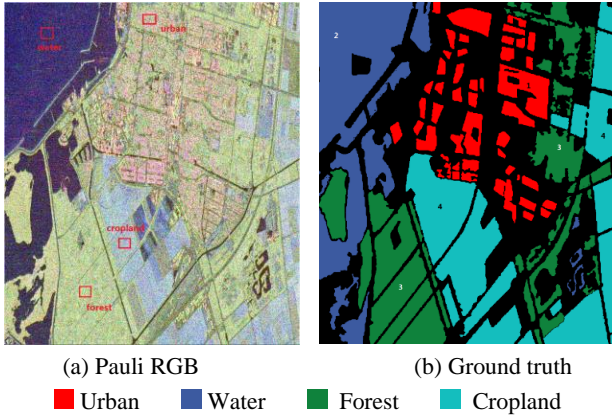

(a) Pauli RGB  (b) Ground truth

■ Urban  ■ Water  ■ Forest  ■ Cropland

Fig.7. (a) Pauli RGB of Flevoland. (b) Ground truth

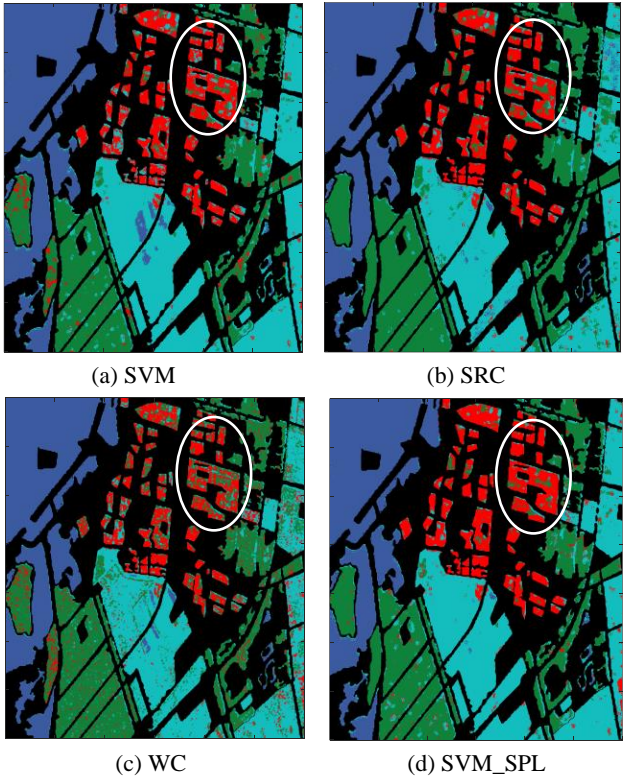

(a) SVM  (b) SRC

(c) WC  (d) SVM_SPL

(e) SVM_SPLNC

Fig.8. (a)~(e) classification results of SVM, SRC, WC, SVM_SPL and SVM_SPLNC

TABLE II
ACCURACIES OF FLEVOLAND DATA SET FROM AIRSAR
AA: AVERAGE ACCURACY; OA: OVERALL ACCURACY

|  | *SVM* | *SRC* | *WC* | *SVM_SPL* | *SVM_SPLNC* |
|---|---|---|---|---|---|
| *Urban* | 0.7169 | 0.7579 | 0.6022 | 0.8270 | **0.8315** |
| *Water* | 0.9695 | 0.9779 | **0.9854** | 0.9732 | 0.9663 |
| *Forest* | 0.8388 | **0.9195** | 0.8479 | 0.8539 | 0.9016 |
| *Cropland* | 0.9425 | 0.8759 | 0.8071 | 0.9512 | **0.9568** |
| ***OA*** | 0.8894 | 0.8978 | 0.8382 | 0.9126 | **0.9177** |

## V. CONCLUSION

In this paper, a novel SVM algorithm based on SPL with neighborhood constraints is proposed for PolSAR image classification. We used this learning mechanism to train the SVM classifier, which learns the easy samples first then gradually involves complex samples into model until the entire training dataset is learned. In addition, a new self-paced regularization term with neighborhood constraints is proposed and implemented. Three measured PolSAR datasets are used to demonstrate the effectiveness of our proposed method. The experimental results indiciate that our proposed method can achieve competitive classification performance on complex scenes with mixed scattering mechanism or on scenes characterized by several types of land cover with similar scattering properties.

## ACKNOWLEDGMENT

This work was supported in part by the National Natural Science Foundation of China (Nos. 61472306, 91438201, 61572383, 61806157), the National Basic Research Program (973 Program) of China (No. 2013CB329402), and the fund for Foreign Scholars in University Research and Teaching Programs (the 111 Project, No. B07048). In addition, the author would like to thank Professor Deyu Meng for help.